# Recognising Affordances in Predicted Futures to Plan with Consideration of Non-canonical Affordance Effects*

Solvi Arnold, Mami Kuroishi, Tadashi Adachi and Kimitoshi Yamazaki

*Abstract*—We propose a novel system for action sequence planning based on a combination of affordance recognition and a neural forward model predicting the effects of affordance execution. By performing affordance recognition on predicted futures, we avoid reliance on explicit affordance effect definitions for multi-step planning. Because the system learns affordance effects from experience data, the system can foresee not just the canonical effects of an affordance, but also situation-specific side-effects. This allows the system to avoid planning failures due to such non-canonical effects, and makes it possible to exploit non-canonical effects for realising a given goal. We evaluate the system in simulation, on a set of test tasks that require consideration of canonical and non-canonical affordance effects.

## I. INTRODUCTION

The concept of affordances, first introduced by Gibson [1] in the field of psychology, has found application in various areas of robotics [2] [3] [4]. Numerous conceptualisations of the affordance concept exist, but affordances can broadly be defined as the action opportunities arising from the combination of an agent's action capabilities with the environment the agent is placed in. Affordances appear to play a central role in structuring high-level behaviour in humans and other animals. Key areas for this work are affordance recognition and affordance-based planning.

Within the area of affordance recognition, the introduction of neural network-based approaches has produced substantial progress over the past years. In [5] and [6], object-recognition-style architectures are applied to the problem of detecting affordances in images. In [7], NNs are used to extract object features that inform object manipulation. CLIPort [8] finds affordances defined as motion start and end points in various scenes on basis of natural language instructions.

Early approaches in planning descend from Situation Calculus [9], and employ grammars to describe action effects and preconditions [10]. The grammatical approach is effective if the task environment can be fully formalised, but hard to adapt to task domains with complex physical effects. This makes rule-based approaches notoriously brittle. Consider the task of stacking objects. We may specify the *canonical* effect of placing one object on another as follows:

$$PLACE\_ON(X,Y) \rightarrow ON(Y,X) \quad (1)$$

This rule may fail to hold for an instance as shown in Fig. 1. Object X can be placed on Y at the arrow marker, but (depending on the relative weights of X and Y) this may not result in X sitting atop Y. We can precondition the rule on stability of the object arrangement, but in practice this implies importing a substantial amount of physics into the precondition check, which quickly makes planning computationally infeasible.

The need to define complex physical preconditions on affordance effects may be avoidable by learning to predict action effects instead. In [11] [12], affordance effects in a block-stacking scenario are learned in a rule-based format for symbolic planning. However, effective use of fine quantitative state features for affordance effect prediction requires that prediction is learned at finer granularity than symbolic rules (learned or given) can provide. In early work in this direction under the nomer of *internal rehearsal* [13], action effects for a traversability affordance are learned using a Gaussian Mixture Model in a goal-specific manner.

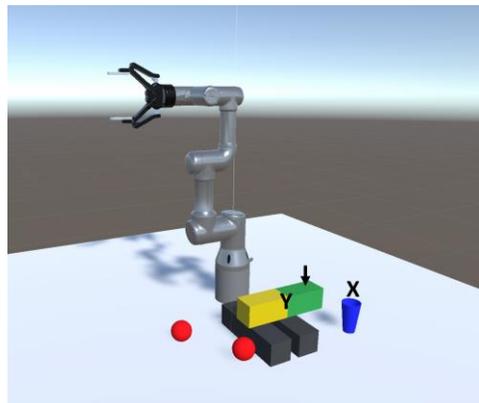

Fig. 1. Snapshot of the simulated environment, showing the virtual UR3 robot and task objects. Letters and arrow marker added.

Outside the context of affordances, more generalised physical prediction has been explored extensively in recent years. In [14], NNs are trained to predict stability (and block trajectories) for block towers specifically, and [15] predicts object motion from real-world images. Models with action input are found in [16] (predictive models for billiard) and [17] (3D rigid body motion). The prediction NN in the present work fits in this line of inquiry.

NN approaches for affordance-based action planning remain scarce, but a notable example is found in [18], which proposes a neural latent dynamics model in which affordance effects include expectations of future affordance availability. However, task-specific Reinforcement Learning is required, and non-canonical effects are not addressed.

The purpose of this work is to combine the affordance concept's potential to structure complex action spaces with the strength of neural prediction for modelling action effects. A key aspect of the approach is that affordance recognition is applied identically on current (actual) states and predicted ("imagined") future states. Hence, future affordances are grounded in the same low-level, sub-symbolic, high-dimensional state representation as current affordances. By carrying sub-symbolic state representations forward through the planning process, we avoid the need for brittle symbolic effect specifications, and avoid planning failures due to hard-to-specify effects and preconditions. Furthermore, the human-interpretable predictions can provide users an intuitive visual explanation of the robot's action choices.

Through simulation experiments, we show that the system produces suitable action sequences for goals provided at runtime, effectively predicting and avoiding non-canonical action effects that would cause failure of logically sensible action choices. We also demonstrate exploitation of such effects to realise other inaccessible goal states.

### A. Canonical and Non-canonical Effects

Symbolic formalisation of affordance effects creates a distinction between *canonical* (formalised) and *non-canonical* (non-formalised) effects. Since we do not formalise effects, our approach makes no inherent distinction between the two. However, the distinction remains conceptually useful. Below we use the distinction informally as follows. Canonical effects are those effects one expects on basis of an affordance's label. Non-canonical effects are any other effects that may accompany execution of an affordance in particular circumstances.

Constraining a planner to formalised effects limits it to action outcomes foreseen and deemed intentional by the rule-designer. However, what may be an undesirable side-effect in one problem setting may be a clever solution in another problem setting. The proposed system can tap this potential.

### B. Contributions

1) We propose a neural architecture that integrates affordance recognition and affordance effect prediction, capable of predicting future affordances through repeat application of scene prediction and affordance recognition.
2) We propose a planning algorithm that uses the neural architecture to plan action sequences that realise positive and negative goal conditions set at run-time.
3) We demonstrate that the system can plan action sequences in consideration of non-canonical affordance effects, avoiding and exploiting such effects as necessary.

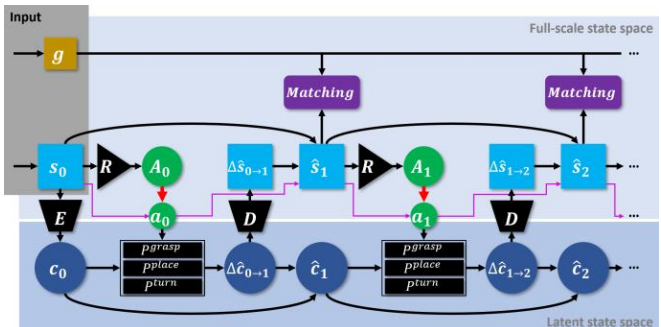

Fig. 2. Global system structure, rolled out for processing affordance sequences of length two. Processing for subsequent affordances is identical to that for the second. Light blue: full-scale state representations, dark blue: latent state representations, green: affordances, black: neural network modules (R: recognition net, E: encoder, P: prediction module, D: decoder), yellow: goal image, purple arrows: affordance pixel data.

## II. SYSTEM ARCHITECTURE

Fig. 2 shows the global structure of our setup. At the core of the system are two neural network modules: an affordance recognition module, and an affordance effect prediction module. We denote states as $s \in S$, let $s_0$ be (the observation of) the current state of the task environment, and use $s_i, i \geq 1$ to refer to possible future states. $A$ denotes the set of affordance types in the robot's repertoire and $n_{aff}$ its size. In our experiments, states will be RGBD images, and $A$ = *{Grasp, Place, Turn}*. Below we explain our setup in detail, starting with our interpretation of the affordance notion.

### A. Affordance Concept

The affordance concept has been defined in numerous ways [2]. Since we purposely exclude definitions of intended effects, we arrive at the following minimalistic definition. In this work, an affordance of type $a \in A$ exists at point $p$ in the problem space IFF the agent is capable of *commencing* an action of type $a$ at $p$. In our experiments, $p$ is a 4D vector consisting of a point in 3D space and a gripper angle, and an affordance of type $a$ exists at $p$ IFF (1) an entity of the type affordances of type $a$ act on exists at the spatial location, and (2) the robot's gripper can reach this location in the orientation specified by the angle, i.e. the gripper can assume the starting pose for the affordance's execution. Given this definition, affordance presence can be evaluated from observation of the current scene alone, without consideration of physical effects beyond commencement. The exclusion of effectiveness at producing a specific outcome from the affordance conceptualisation echoes [18].

### B. Affordance Recognition

Some existing affordance recognition methods perform recognition at pixel-level granularity [5], alongside semantic recognition of the object. However, for integration in our planning architecture, it is desirable to recognise affordances in a format that can be interpreted as a physical action directly, without further processing. This approach is consistent with the finding that visual processing for object-directed action in humans occurs separate from and independent of semantic recognition [19]. Our recognition result format consists of the following values: (1) the affordance type, (2) a target point in 3D space, (3) a gripper angle. For the *grasp* and *turn* affordances in our experiments, the target point indicates the centre of the object to be grasped or turned, and for the *place* affordance, it indicates the point (on some surface) that the currently held object is to be placed at. Affordances may have additional parameters determining details of their execution. In the current repertoire, the *turn* affordance has a second angle parameter defining the turn amount and direction. However, because we define presence of an affordance in terms of *possibility to commence*, such parameters are not part of the recognition format.

We additionally include a value indicating the affordance's symmetry w.r.t. the gripper angle. This value indicates whether the affordance is qualitatively distinct from other valid affordances that differ only by the gripper angle. Consider grasping or placing a rotationally symmetrical object. The object will allow multiple (potentially infinitely many) valid gripper angles, but these angles produce qualitatively equivalent results. Recognising such affordance equivalence allows us to improve efficiency of the planning process.

Our recognition format (a point in space plus some parameters) resembles that of the YOLO [20] family of object recognition architectures (a point in space plus bounding box dimensions). We adapted the ScaledYOLOv4 architecture [21] to perform affordance recognition instead of object recognition. In the interest of space, we limit our discussion of the architecture to our main modifications.

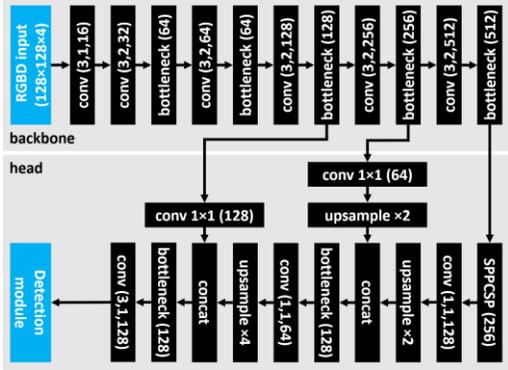

Fig. 3. ScaledYOLOv4 Network configuration used for affordance recognition. Numbers in brackets indicate kernel size, stride, and number of output channels for convolution layers, and output channels for other modules.

We change the input format from RGB to RGBD. We drop the bounding box width and height outputs, replacing them with the affordance angle. ScaledYOLOv4 originally uses bounding box anchors to allow for multiple different detections at nearby locations. We adapt the anchor logic to allow for multiple affordances with different angles at nearby locations, setting anchors at 90° intervals. Additionally, we let detections include a z-coordinate and a symmetry value. Training losses are adapted accordingly, replacing the bounding box overlap loss with MSE losses for angles, z-coordinates, and symmetry values. We repurpose YOLO's 'objectness' as detection confidence. The architecture of the backbone and head networks are reconfigured as shown in Fig. 3. Since the visual data in our experiments is relatively simple and the number of affordance types small, we reduce the bandwidth of all layers of the default ScaledYOLOv4 network by a factor two. The default network performs detections separately on basis of the output of three layers in the network, with the resolution of the detection grid decreasing in order. Since affordances in our experimental setup can exist in close proximity, a relatively fine detection grid is necessary. The coarser layers were found to produce spurious detections (apparent averages over the affordances within a grid cell), so we constrain input to the detection module to a single layer. The original architecture prunes detections by objectness, and subsequently applies non-max suppression (NMS) of overly similar results on basis of bounding box overlap. We similarly prune by confidence, but for lack of bounding boxes, we filter detections of the same affordance type by proximity in state-angle space. When multiple detections are in overly close proximity, we retain the detection with the highest confidence. The prediction network is implemented by modification of the PyTorch [22] implementation of ScaledYOLOv4.

### C. Affordance Effect Prediction

The prediction pathway is based on the EM*D architecture [23] [24]. The present work extends this architecture with differential prediction and affordance-specific prediction modules. The motivation for introducing differential prediction is that complex scenes may contain any number of elements that remain unaffected by execution of a given affordance. Reconstructing these elements in predicted states is unnecessary, increases the difficulty of the prediction problem, and degrades prediction quality. Instead, we let the system indicate where changes occurred, and fill in the changed parts of the state. This allows for unchanged elements to be retained in their original fidelity.

The prediction pathway consists of Encoder ($E$), Decoder ($D$), and $n_{aff}$ prediction modules $P^a, a \in A$. $E$ maps input state $s_0$ (an RGBD image in our implementation) to its latent representation $c_0$. $P^a$ takes a (predicted or actual) latent representation $c_i$ and a parametrisation $a_i$ of affordance $a$ as input and generates prediction $\Delta \hat{c}_{i \to i+1}$ of the difference $\Delta c_{i \to i+1}$ between the latent representation $c_{i+1}$ of the state $s_{i+1}$ resulting from performing $a_i$ in $s_i$ and the latent representation $c_i$ of state $s_i$. Latent prediction $\hat{c}_{i+1}$ is obtained from $c_i$ and $\Delta \hat{c}_{i \to i+1}$ through simple summing:

$$\hat{c}_{i+1} = c_i + \Delta \hat{c}_{i \to i+1} \qquad (2)$$

$D$ maps a latent difference representation $\Delta \hat{c}_{i \to i+1}$ to a full-scale representation $\Delta \hat{s}_{i \to i+1}$ of the state difference described therein.

The format of the full-scale difference representation is as follows. For an n-channel (e.g. 4 for RGBD images) state format, we use an (n+1)-channel difference descriptor, with the additional channel encoding a mask that determines where the other channels should overwrite the original state. Full-scale difference representations are thus applied to state representations as follows:

$$\hat{s}_{i+1} = s_{i \to i+1}^{c+1} * s_{i \to i+1}^{1:c} + (1 - s_{i \to i+1}^{c+1}) * s_i \qquad (3)$$

Where c is the channel count of the state representation, and superscripts indicate channel selection.

Network architectures in the prediction pathway are as follows. E consists of 2 convolutional layers with kernel size 3, stride 2, and output channel count 8, followed by 3 dense layers of 8192, 4096, and 128 neurons. D consists of 3 dense layers followed by 2 up-convolutional layers, with channel and neuron counts mirroring E. Following [25], we use 2x nearest neighbour upscaling to increase resolution between up-convolutional layers.

Each $P^a$ module duplicates the architecture shown in Fig. 4. The $P^a$ module used to process a given action input is determined by the affordance type value in that input. Hence actions are not simply passive input signals, but actively determine the course of signal propagation through the prediction pathway. The remaining values are input to the selected module. The 5th action input value is only used for the *turn* affordance (for other affordances it is set to 0). The last four inputs contain the RGBD values of a single pixel from the image representing the state in which the affordance is to be performed. For the initial state, this image is already part of the input. For subsequent states, we use the prediction generated by the network. Hence, no additional external information is required to provide the pixel input. We project the 3D position of the affordance to pixel coordinates, and retrieve the RGBD values from the corresponding pixel.

As shown in Fig. 4, the $P^a$ module has I/O blocks for a memory trace. The memory trace is passed between subsequent

$P^a$ modules in the pathway. When input is received from $E$, memory trace input is a zero vector.

All networks in the prediction pathway use the hyperbolic tangent activation function at all layers, except for the output layer of D, which simply clamps output activation to the [0,1] range. The prediction pathway is implemented in JAX [26].

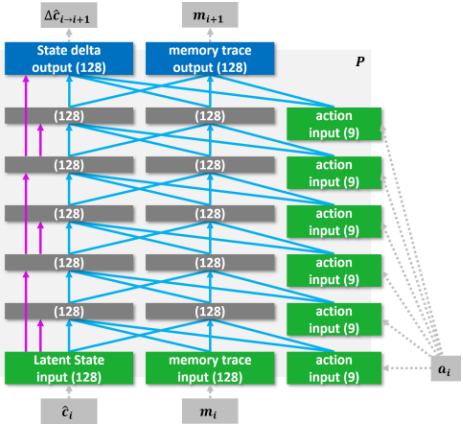

Fig. 4. *P* module architecture. Each block is a set of neurons. Numbers in brackets indicate neuron counts. Green: input, blue: output, dark grey: hidden. Action input is identical across layers. Blue lines indicate full connectivity between sets. Purple lines indicate 1-to-1 copying of activation values.

### D. Planning & Execution Logic

Action sequences are planned as follows. We define the planning goal using an RGBD or RGB goal image depicting the objective to be achieved. By using a goal image of size smaller than the state image resolution, we can define objectives without having to specify the full goal state. For example, we can specify the goal of placing a cup on a block without specifying positions for other objects in the scene.

We expand a search tree of possible futures from the current state by repeated recognition and prediction, as illustrated in Fig. 5. Each edge of the search tree corresponds to a fully parametrised affordance, and each node corresponds to a state. Given a state, recognition thus provides the edges extending from that state. To predict child state *s* for a given edge, we run prediction from the current state (root node of the search tree), with the chain of affordances leading up to *s* as action input[1].

Affordance parameters aside from position, gripper angle, and symmetry (i.e. parameters not affecting existence of the affordance at a given location) are filled in between recognition and prediction. Our experiments feature one such parameter: the turn angle parameter of the *turn* affordance. Such affordances are duplicated with different parametrisations. For the turn affordance, we parametrise the turn angle to ±90°. The recognition process may return multiple affordances that differ only in their gripper angle, and are rotationally symmetrical (i.e. symmetry parameter value > 0.5). For any set of such affordances, we only process the affordance with the highest confidence and discard the rest. This avoids unnecessary branching of the search tree.

Each predicted state is evaluated against the goal image (*matching* in Fig. 2) with a simple sliding window method. For a state image resolution of $w^s \times h^s$, we define a goal image as an image of resolution $w^g \times h^g$, $w^g \leq w^s$, $h^g \leq h^s$. We can then define the planning problem as follows.

$$plan = \underset{a_{0:n-1}}{\operatorname{argmin}} L(g, Pr(s_0, a_{0:n-1})), n < n_{max} \quad (4)$$

$$L(g, \hat{s}) = \min_{0<x<w^s-w^g, 0<y<h^s-h^g} MSE(g, \hat{s}_{x:x+w^g, y:y+h^g}) \quad (5)$$

Where $Pr(s_0, a_{0:n-1})$ applies the prediction pathway with state $s_0$ and parametrised affordance sequence $a_{0:n-1}$, $n_{max}$ is the maximum sequence length to consider, and $w^s \times h^s$ is the state image resolution. In other words, the loss for a given plan (i.e. sequence of parametrised affordances) is the smallest MSE loss between the goal image and any patch of size $w^g \times h^g$ in the prediction generated by that plan. Typically, we return the plan with the lowest residual loss. However, we can also define "negative" goals, i.e. objectives to avoid. By selecting the affordance sequence producing the *highest* residual loss (i.e. replacing *argmin* with *argmax* in Eq. 4), we find plans for avoiding or eliminating the situation expressed by the goal image.

Each process (recognition, prediction, evaluation) can be parallelised per depth level of the search tree by means of batch processing.

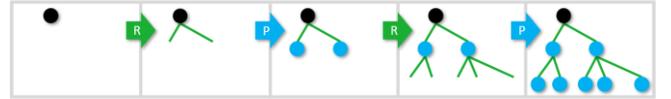

Fig. 5. Search tree expansion through repeated affordance recognition and state prediction. Black dot: current state, edges: detected affordances, blue dots: predicted states, R: recognition, P: Prediction.

## III. EXPERIMENTAL SETUP

### A. Task Space

Fig. 1 shows a snapshot of our task scene. The scene is implemented in Unity [27]. The agent controls a virtual UR3 (Universal Robots) robot placed on a table. Motion planning is performed using MoveIt via a ROS connection, as provided by the Unity Robotics Hub [28]. The scene contains 0-3 cups, 0-3 balls, 0-2 black "support" blocks, and one long coloured block, arranged in a rectangular area in front of the robot.

We define the set of affordances *A* as *{Grasp, Place, Turn}*. *Grasp* affordances exist at the centre point of each object, except for the support blocks. *Place* affordances exist at the centre of the top surface of each support block, and on 3 positions on the top surface of the coloured block. A *turn* affordance exists at the centre of the coloured block. The turn affordance turns the block 90° clockwise or counter-clockwise in the XY plane. After execution of an affordance, the robot returns to a default pose with its gripper pointing upward. After execution of a grasp affordance, the gripper holds the grasped object.

Following our *possibility to commence* affordance existence criterion, the existence of the above affordances is conditioned on (1) the current state of the gripper (*grasp* and *turn* affordances only exist if the gripper is empty, and *place* affordances only exist when the gripper is holding something), and (2) the starting pose being collision-free and within reach of the robot. No consideration is given to whether execution would produce the affordance's canonical result.

---

[1] Prediction could also be performed as single-step prediction from the state's parent state, but (for states removed from the root state by more than one affordance) this increases the number of encoding and decoding steps, which degrades prediction quality.

Various interactions between objects occur in this setup. Turning the block can cause nearby cups to fall over, and nearby balls to roll away. It will also cause any balls on top of the block to roll away, but a cup on top of the block will stay put and turn along with the block. Lifting the block when there is another object on top of it will cause that object to fall off. Placing an object on top of the coloured block when the coloured block is on top of a support block may produce an unstable situation, causing the block to tilt over and the object to fall off or roll away. Hence while the scene is simple, it produces a variety of non-canonical effects.

State images are captured from a top-down perspective to minimise occlusion, and rescaled to 128×128 resolution. The robot itself is visible in state images, and consequently states encode any object the gripper is holding.

### B. Data Generation

We generate data for training the recognition and prediction networks as follows. We randomise the arrangement of objects in the scene (position of each object, and number of instances for each object except the coloured block). We then perform a simple affordance detection routine. Each affordance location is represented in the simulation environment as an invisible marker attached to its host object. We check whether the marker is within reach of the robot, and if so, check whether the affordance's commencement pose is accessible by placing a gripper-shaped object at the marker position and checking for collisions. If the gripper pose is collision-free, we record the affordance. For *grasp* and *turn* affordances on the block, the gripper angle is aligned to the block's orientation. For *grasp* affordances on radially symmetrical objects, as well as all *place* affordances, we perform the accessibility check at 90° intervals of the gripper angles. For *place* affordances, we let the gripper-shaped object hold a copy of the currently held object when checking for collisions. We record all found affordances as the *affordance list* for the state. We then randomly select and perform one affordance from the list, obtaining a new state. We repeat this process four times for each data sequence, or until no affordances remain in the scene. Each *(state, affordance list)* pair provides one example for the recognition network, while each sequence of states and executed affordances provides an example for the prediction network. We collected a dataset of 25850 sequences containing 97377 affordance executions.

### C. Training

The Recognition network was trained using the standard ScaledYOLOv4 training procedure for 50 epochs. For data augmentation, we apply random small translations to states, shifting affordance positions accordingly. Test and validation sets consist of 100 examples each.

The prediction pathway $(E, P, D)$ was trained for 1m batches of 32 sequences each, using the SignSGD weight update rule [29] with an initial learning rate of $5 \times 10^{-5}$. We evaluate performance on the validation set once every 10000 batches, and if performance does not improve for 5 subsequent evaluations, we reduce the training rate by a factor 2. Sequence length is varied per training batch from 1 to 4 affordances, and starting points within example sequences are selected randomly (where possible). Note that the sequence of $P$ modules varies per example (being determined by the sequence of affordance types in the example). Training with various $P$ module orderings ensures that the learned latent representation format is compatible between $P$ modules. For data augmentation we use translations, mirroring, noise on action input values, and 180 degree turns of the affordance angle in cases where the angle is irrelevant. Test and validation sets consist of 500 examples each.

## IV. EVALUATION

### A. Evaluation – Recognition

We evaluate the recognition module in terms of recall and spurious detections for a given cut-off threshold on detection confidence. Recall is satisfied for a given affordance if a detection of the correct type is produced within 2.5cm from the ground truth position, with an angle error of <5°, which is sufficiently precise for affordances to be executable in the simulation environment. For scale, the long side of each block measures 28cm. Detections outside the above ranges from a ground truth affordance are considered spurious. The confidence cut-off (0.9) is tuned on the validation set. We measure performance on the test set and on 100 examples from the training set. Quantitative results are shown in Table 1, and detection examples are shown in Fig. 6.

TABLE 1   AFFORDANCE RECOGNITION ACCURACY

|  | Grasp | Place | Turn | All | Spurious |
|---|---|---|---|---|---|
| **Test** | 0.988 (324/328) | 0.941 (384/408) | 0.976 (82/84) | 0.963 (790/820) | 0.20/image |
| **Train** | 0.987 (298/302) | 0.997 (596/598) | 0.931 (54/58) | 0.990 (948/958) | 0.10/image |

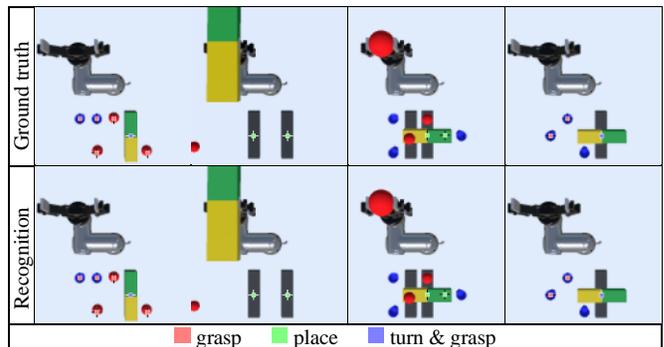

Fig. 6. Examples of affordance recognition. Lines extending from markers indicate grasp angles, with the colour of the line indicating whether the grasp angle affects the outcome (white) or not (black). Note that accessibility of specific grasp angles is affected by surrounding objects. Grasp and turn affordances are detected when the gripper is empty, and placement affordances are detected when the gripper holds an object.

Note that the robot state is visible in the state image, and recognition incorporates this information correctly. Detection errors primarily stem from 1) misjudgement of object accessibility in crowded arrangements, and 2) failure to mark grasp affordances on toppled cups as asymmetrical (most instances of graspable cups in the dataset are in upright position).

### B. Evaluation – Prediction

For quantitative evaluation of the prediction pathway, we perform prediction for all sequences in the test set and calculate pixel value accuracy. Given our differential prediction setup, substantial areas of the state image are copied from one state to the next. To evaluate accuracy w.r.t. changed state areas specifically, we include pixel value accuracy for areas that have changed from the preceding state (in the ground truth sequence) as a secondary evaluation metric ("changed area accuracy"). Absolute mean RGBD pixel value error (range [0,1]) is given in Table 2. We observe that the mean error for changed areas is ≤0.04 for all sequence lengths when the affordance pixel is included in the input, with no significant overfitting.

TABLE 2     PREDICTION ACCURACY

| Set | Step | With affordance pixel | | Without affordance pixel | |
|---|---|---|---|---|---|
| | | All-area accuracy | Changed area acc. | All-area accuracy | Changed area acc. |
| Test | 1 | **.0024 (.0013)** | **.034 (.017)** | .0025 (.0015) | .035 (.020) |
| | 2 | **.0031 (.0028)** | .034 (.026) | .0032 (.0028) | .034 (.026) |
| | 3 | **.0048 (.0059)** | **.039 (.030)** | .0050 (.0060) | .041 (.034) |
| | 4 | **.0046 (.0037)** | **.035 (.025)** | .0048 (.0039) | .037 (.030) |
| | All | **.0037 (.0039)** | **.035 (.025)** | .0038 (.0040) | .037 (.028) |
| Train | 1 | .0025 (.0016) | .035 (.017) | **.0024 (.0025)** | **.034 (.016)** |
| | 2 | .0030 (.0028) | .032 (.020) | .0030 (.0027) | **.032 (.019)** |
| | 3 | .0041 (.0028) | .036 (.025) | **.0040 (.0027)** | **.036 (.024)** |
| | 4 | .0047 (.0041) | **.035 (.023)** | .0047 (.0041) | .036 (.025) |
| | All | .0035 (.0030) | .035 (.021) | .0035 (.0030) | **.034 (.021)** |

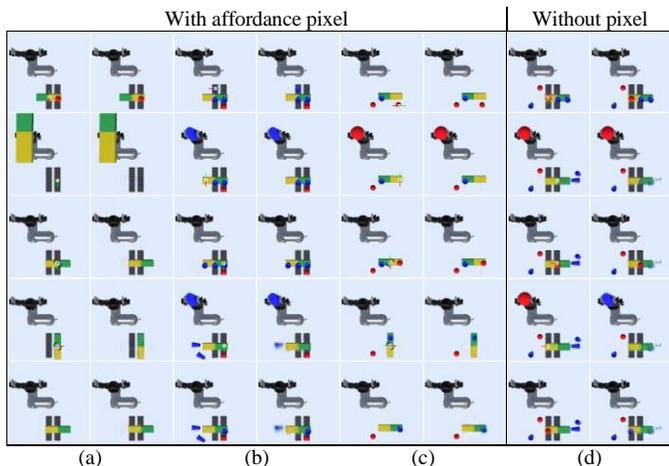

(a)      (b)      (c)      (d)

Fig. 7. Example predictions (test set). In each example, left column shows ground truth state sequence along with executed affordances, and right column shows initial and predicted states. Red and green lines on markers correspond to gripper fingers. For *turn* affordances, black quarter circles indicate the turn direction. a. Moving the block around. Note that the ball falls and rolls away. b. Cup balancing. The first placement is stable because of counterbalancing. Lifting the cup on the right breaks the balance, causing the block to topple and the remaining cups to fall off. The resulting pose and location of the left cup is predicted approximately (blue blur). The trajectory of the middle cup proves too unpredictable. c. Turning a block with objects on it. Note that the ball rolls away when the block is turned, while the cup stays put. d. Example of object confusion in a crowded scene by a network trained without the affordance pixel input.

Note that the dataset includes inherently unpredictable outcomes. Effects of uncontrolled kinetic interactions between objects (e.g. unstably placed objects falling over or rolling away) are not exactly predictable. Consequently, exact prediction is impossible. However, perfect prediction is not necessary for effective planning. For example, as long as the prediction for an unstable placement captures the fact that the placed object will not remain in the location where it was placed, it will suffice to avoid planning that placement. As illustrated by Fig. 7, predictable effects (panel a, c) are well predicted, while unpredictable effects (panel b) produce messier results, as expected. Note that when objects are placed on the coloured block in an unstable manner as in Fig. 7b, the block tilts and the object falls off, but once the object no longer weighs down the block, the block usually tilts back into a horizontal orientation. This behaviour is physically correct.

---

[2] https://youtu.be/4naJ5IghHcg

Affordance pixel input has no appreciable effect on the visual quality of predictions, but nets trained without the pixel input occasionally confuse grasped objects in complex scenes, as seen in Fig. 7d. This can easily cause planning failures. So, while the quantitative accuracy difference is small, nets trained with affordance pixel input are more suitable for planning.

### C. Evaluation – Planning

For evaluation of the integrated system, we prepared a set of 10 test tasks, each designed to require some modicum of insight in the scene's dynamics. Tasks and generated solutions are shown in Fig. 8. Note that in each plan, states after the first are predictions generated by the prediction network. Footage of the robot executing solutions can be found in the video accompanying this paper[2]. Below we briefly describe the challenge posed by each task.

1) This task requires consideration of action order, because the red ball occupies the goal position for the blue cup.
2) The cup is initially inaccessible, requiring that the coloured block is moved out of the way first, and restored to its original pose after placing the cup.
3) Placing the cup on the coloured block directly would be unstable, causing the block to fall. Moving the block allows stable placement of the cup.[3]
4) This task requires counterbalancing the blue cup with the red ball. Hence, the ball must be placed in its target location on the coloured block before the cup.
5) Same concept as task 4, but changing the location of one support block, so that now the cup must be placed on the coloured block before the ball.
6) Turning the block would knock over the cup. Solving the task requires understanding that a cup placed on a block turns along with that block.
7) The red ball should be moved out of the area in front of the block, but the gripper cannot reach the ball due to the placement of the cup. The system must exploit the fact that a turning block pushes away balls in its path, and pick the correct turn direction to achieve this effect.
8) Negative goal: eliminate the ball. Solved here by creating an unstable block configuration to use as a slope for the ball to roll off of.
9) Negative goal: eliminate the cup from the scene. The affordance repertoire provides no straight-forward way of removing objects from the scene. Solved here by first placing the cup on the block, and then lifting the block, causing the cup to fly off.
10) Negative goal: eliminate the ball. The ball is inaccessible due to the cup placement. However, the robot can hide it from its own view by placing the coloured block over it.

Tasks 1, 2 are solved using canonical affordance effects plus consideration of how execution of those affordances determines which affordances become available in future states. Tasks 3-10 involve consideration of non-canonical affordance effects. Tasks 4 and 5 only differ in the location of one support block, but require different action orders. The system responds accordingly, demonstrating that it considers the stability of intermediate states. Tasks 8-10 use negative goals. For these cases we use small goal patches, and ignore the depth channel in the plan loss calculation. This is to eliminate the trivial solution where the object is merely lifted up (lifting an object up changes its visual size and depth values). The system comes up with various ways of eliminating objects from the scene, exploiting non-canonical affordance effects to satisfy the goal condition.

---

[3] The goal image in this task is prepared with the block in its original position, to avoid leaking the solution through perspective information

| | #1 (4 steps) | #2 (4 steps) | #3 (4 steps) | #4 (4 steps) | #5 (4 steps) | #6 (4 steps) | #7 (2 steps) | #8 (4 steps) | #9 (4 steps) | #10 (2 steps) |
|---|---|---|---|---|---|---|---|---|---|---|
| Goal | | | | | | | | NOT (no D) | NOT (no D) | NOT (no D) |
| Start | | | | | | | | | | |
| Plan | | | | | | | | | | |
| Result | | | | | | | | | | |

Fig. 8. Examples of generated plans and their execution results. Task concepts are explained in Section 4C. Goal image are presented in RGBD or RGB format (the depth channel is omitted in negative goals to avoid solutions that merely lift up the object to be eliminated). Plans consist of affordance sequences. We draw the affordance sequence on a state sequence consisting of the initial state and network-generated predictions of the subsequent states for the plan. States are in RGBD format, but the D channel is omitted for space consideration.

## V. DISCUSSION & FUTURE WORK

Our results demonstrate affordance-structured reasoning with consideration of non-canonical effects. The system effectively avoids non-canonical effects that would obstruct access to the goal state, and allows non-canonical effects to be actively exploited to reach goals that are otherwise inaccessible.

The system still has a number of significant limitations. We currently operate on discrete affordances. However, many real-world affordances have continuous components. Place affordances would be better represented as regions instead of points in space, and parameters such as the turn angle of the turn affordance should be generalised to continuous values. We will extend the affordance format to cover continuous elements, and implement run-time optimisation of the parameters of selected affordances using a back-propagation strategy as employed in previous work on the EM*D architecture [24].

A strength of the system is that it takes goals at run-time. However, goals are currently provided in image form, which is impractical in practice. To make use of the flexibility of run-time goal specification in practical use scenarios, a more convenient style of defining goals will be desirable. One strategy may be to replace the evaluation function of Eq. (4) with a neural network that quantifies agreement between predicted outcomes and sentences describing goals in natural language, using e.g. the CLIP architecture [30].

Another open issue is how to handle actions with stochastic outcomes. Uncertainty is expressed to some extent in the form of blurred prediction for objects whose position cannot be predicted with certainty (cf. Fig. 7b), but a more principled approach is desirable. Elsewhere [*24*], we perform prediction probabilistically, quantifying prediction uncertainty. We are exploring affordance recognition on probabilistic predictions, to enable reasoning through stochastic action outcomes.

While the present system does not rely on symbolic affordance effect rules, we do believe a hybrid approach can have significant advantages. For cases where symbolic planning is effective, it is more efficient. Where symbolic planning is tripped up by non-canonical effects, sub-symbolic prediction can provide an advance reality check to weed out invalid plans. When no plan can be found symbolically, purely non-symbolic planning can provide a fallback strategy to find plans that rely on non-canonical effects. If rules are learned (as in [*11*] [*12*]), non-symbolic planning may be exploited to gather experience, allowing for incremental "canonicalisation" of such effects. Exploration of such hybridisation remains as future work.

Finally, in our experimental setup, the number of affordances per state is limited, and we can feasibly expand the search tree to reasonable depth through parallelisation. For more affordance-rich environments or longer sequences, we will need heuristics to constrain which branches are expanded. Here too, hybridisation could likely be beneficial.